\documentclass[12pt,journal,compsoc,onecolumn,draftclsnofoot,]{IEEEtran}

\makeatletter
\def\thickhline{%
	\noalign{\ifnum0=`}\fi\hrule \@height \thickarrayrulewidth \futurelet
	\reserved@a\@xthickhline}
\def\@xthickhline{\ifx\reserved@a\thickhline
	\vskip\doublerulesep
	\vskip-\thickarrayrulewidth
	\fi
	\ifnum0=`{\fi}}
\makeatother

\newlength{\thickarrayrulewidth}
\usepackage{epigraph}
\usepackage{longtable}
\usepackage{setspace}

\usepackage{textcomp}

\usepackage{amssymb}

\usepackage[hidelinks]{hyperref}
\hypersetup{
	colorlinks=true,
	linkcolor=black,
	filecolor=black,
	citecolor=black,
	urlcolor=blue,
}

\usepackage{multirow}

\usepackage{color}
\usepackage[table, dvipsnames]{xcolor}
\definecolor{PyOrange}{RGB}{255, 201, 14}
\definecolor{PyBlue}{RGB}{112, 146, 190}

\definecolor{WordGreen}{RGB}{100, 136, 40}
\definecolor{WordDarkGrey}{RGB}{82, 82, 82}
\definecolor{WordRed}{RGB}{192, 80, 77}
\definecolor{WordBlue}{RGB}{0, 122, 192}

\definecolor{WordLightBlue}{RGB}{218, 238, 243}
\definecolor{WordLightGreen}{RGB}{234, 241, 221}

\definecolor{WordFillGreen}{RGB}{194, 214, 155}
\definecolor{WordFillRed}{RGB}{252, 214, 182}
\definecolor{WordFillGray}{RGB}{217, 217, 217}

\UseRawInputEncoding
\usepackage{times}
\usepackage{epsfig}
\usepackage{amssymb}
\usepackage[ruled,vlined,linesnumbered]{algorithm2e}

\usepackage{comment}
\usepackage{array, makecell} 
\usepackage{lettrine}
\usepackage{url} 
\usepackage{lscape}
\usepackage{cuted}
\usepackage{tabularx}
\usepackage{colortbl}
\usepackage{hhline}
\usepackage{vcell}
\usepackage{longtable}
\usepackage{vcell}
\usepackage{rotating}
\usepackage{pdflscape}
\usepackage{sidecap}
\usepackage{neuralnetwork}
\usepackage[export]{adjustbox} 
\usepackage{acronym}
\acrodef{FCN}[FCN]{Fully Convolutional Network}
\acrodef{GAME}[GAME]{Grid Average Mean Absolute Error}
\acrodef{DL}[DL]{Deep Learning}
\acrodef{DNN}[DNN]{Deep Neural Network}
\acrodef{ML}[ML]{Machine Learning}
\acrodef{CV}[CV]{Computer Vision}
\acrodef{AI}[AI]{Artificial Intelligence}
\acrodef{CNN}[CNN]{Convolutional Neural Network}
\acrodef{RNN}[RNN]{Recurrent Neural Network}
\acrodef{GAN}[GAN]{Generative Adversarial Network}
\acrodef{JCU}[JCU]{James Cook University}
\acrodef{MAE}[MAE]{Mean Average Error}
\acrodef{MAP}[mAP]{Mean Average Precision}
\acrodef{CA}[CA]{Classification Accuracy}
\acrodef{LCFCN}[LCFCN]{Localization-based Counting loss Fully Convolutional Network}
\acrodef{IoT}[IoT]{Internet of Things}
\acrodef{MLP}[MLP]{Multi-Layer Perceptrons}
\def\x{{\mathbf x}}				
\def\y{{\mathbf y}}				
\def\W{{\mathbf W}}				
\def\bR{\mathbb{R}}				
\usepackage{xspace}

\usepackage[capitalize]{cleveref}
\crefname{section}{Sec.}{Section}
\usepackage{booktabs}
\usepackage{arydshln}
\usepackage{listings}
\definecolor{codegreen}{rgb}{0,0.6,0}
\definecolor{codegray}{rgb}{0.5,0.5,0.5}
\definecolor{codepurple}{rgb}{0.58,0,0.82}
\definecolor{backcolour}{rgb}{0.95,0.95,0.92}
\definecolor{cleacolorr}{rgb}{1,1,1}

\lstdefinestyle{mystyle}{
    backgroundcolor=\color{cleacolorr},   
    commentstyle=\color{codegreen},
    keywordstyle=\color{magenta},
    numberstyle=\tiny\color{codegray},
    stringstyle=\color{codepurple},
    basicstyle=\ttfamily\footnotesize,
    breakatwhitespace=false,         
    breaklines=true,                 
    captionpos=b,                    
    keepspaces=true,                 
    numbers=left,                    
    numbersep=5pt,                  
    showspaces=false,                
    showstringspaces=false,
    showtabs=false,                  
    tabsize=2
}

\usepackage[most]{tcolorbox}
\newtcolorbox[auto counter]{pabox}[2][]{%
colback=blue!5!white,colframe=blue!75!black,fonttitle=\bfseries,
title=Box~\thetcbcounter: #2,#1}

\newcommand{\MyPaperTitle}{Prawn Morphometrics and Weight Estimation from Images using Deep Learning for Landmark Localization}

\usepackage{tikz}

\ifCLASSOPTIONcompsoc
  \usepackage[caption=false,font=normalsize,labelfont=sf,textfont=sf]{subfig}
\else
  \usepackage[caption=false,font=footnotesize]{subfig}
\fi

\usepackage[numbers,sort&compress]{natbib}

\usepackage{graphicx}
\graphicspath{{./Graphics/}}
\DeclareGraphicsExtensions{.pdf,.jpeg,.png,.jpg}

\usepackage{amsmath}

\usepackage{array}

\usepackage{url}
\usepackage[switch]{lineno}

\begin{document}

\title{\MyPaperTitle}


 \author{
    \IEEEauthorblockN{Alzayat Saleh\IEEEauthorrefmark{1}, 
    Md Mehedi Hasan\IEEEauthorrefmark{2},    Herman W Raadsma\IEEEauthorrefmark{2},    Mehar S Khatkar\IEEEauthorrefmark{2},
    Dean R Jerry\IEEEauthorrefmark{1}, 
    and Mostafa~Rahimi~Azghadi\IEEEauthorrefmark{1}}
    
    \IEEEauthorblockA{\IEEEauthorrefmark{1}College of Science and Engineering, James Cook University, Townsville, QLD, Australia}
    
    \IEEEauthorblockA{\IEEEauthorrefmark{2}Sydney School of Veterinary Science, Faculty of Science, The University of Sydney, Camden, NSW, Australia}
    
}
 
\maketitle

\begin{abstract}
Accurate weight estimation and morphometric analyses are useful in aquaculture for optimizing feeding, predicting harvest yields, identifying desirable traits for selective breeding, grading processes, and monitoring the health status of production animals. However, the collection of phenotypic data through traditional manual approaches at industrial scales and in real-time is time-consuming, labour-intensive, and prone to errors. Digital imaging of individuals and subsequent training of prediction models using Deep Learning (DL) has the potential to rapidly and accurately acquire phenotypic data from aquaculture species.  In this study, we applied a novel DL approach to automate weight estimation and morphometric analysis using the black tiger prawn (Penaeus monodon) as a model crustacean. The DL approach comprises two main components:   a feature extraction module that efficiently combines low-level and high-level features using the Kronecker product operation;  followed by  a landmark localization module that then uses these features to predict the coordinates of key morphological points (landmarks) on the prawn body. Once these landmarks were extracted, weight was estimated using a weight regression module  based on the extracted landmarks using a fully connected network. For morphometric analyses, we utilized the detected landmarks to derive five important prawn traits. Principal Component Analysis (PCA) was also used to identify landmark-derived distances, which were found to be highly correlated with shape features such as  body length, and width. We evaluated our approach on a large dataset of 8164 images of the Black tiger prawn (Penaeus monodon) collected from Australian farms. Our experimental results demonstrate that the novel DL approach outperforms existing DL methods in terms of accuracy, robustness, and efficiency.
\end{abstract}

\ifCLASSOPTIONpeerreview
\else
\fi

\section{Introduction}\label{secintro}
The farming of marine prawns (shrimp) is one of the most important and largest aquaculture production sectors globally \cite{Sang2020GenotypeMonodon}. As in any animal production sector, the acquisition of industrial-scale data on weight and other commercially relevant phenotypic traits is important, as this data can improve yields and economic efficiency through informing pond management, feeding, grading and selective breeding processes \cite{Boyd2022PerspectivesReview, Mitteroecker2009AdvancesMorphometrics}.  However, the current collection of this data is manual using traditional weight and morphometric analyses that are often invasive, time-consuming, labour-intensive, and prone to human error. Therefore, there is a need for the development of automated, fast, and accurate methods for weight estimation and associated morphometric analyses.

Computer vision and image analysis enabled by Deep Learning (DL) have emerged as promising techniques for solving various problems in the Internet of Underwater Things (IoUT) \cite{Jahanbakht2021} and equally in aquaculture \cite{Konovalov2019, saleh2022a}. In particular, image analysis can be used to identify prawn species, detect prawns in images, measure prawn length, and estimate prawn weight \cite{Setiawan2022ShrimpApproach}. However, existing methods have some limitations, such as requiring high-quality images with uniform backgrounds, relying on hand-crafted features that may not capture complex variations, or using simple regression models that may not generalize well to different conditions \cite{Vo2021OverviewVision}. Moreover, most of these methods do not consider the morphological characteristics of prawns that affect their weight distribution.

In this paper, we propose a novel Deep Learning \cite{Laradji2021, Saleh2020a} approach for automated morphometric analyses and weight estimation of prawns from digital images. Our approach consists of two main components: a Kronecker product-based feature extraction module (KPFEM), and a landmark localization module (LLM). The KPFEM uses the Kronecker product operation \cite{Devi2021KroneckerProduct} to combine low-level and high-level features from different convolutional layers efficiently. The LLM predicts the coordinates of key points on the prawn body using a localization network. For weight estimations, we have designed a weight regression module (WRM) that works based on the extracted landmarks using a fully connected network. We also use the landmarks generated by the LLM component to perform morphometric analysis. To the best of our knowledge, this is the first work that applies the Kronecker product operation for feature extraction in morphometrics analysis. Moreover, this is the first work that uses a localization   network for landmark detection in prawn images.

The main contributions of our paper are as follows:

\begin{enumerate}
    \item We introduce a novel feature extraction method based on Kronecker product that can capture rich semantic information from different scales from the prawn image while offering advantages such as reduced model parameters that make it well-suited for resource-constrained devices in aquaculture settings, and improved performance compared to traditional CNNs.
    \item We apply a deep network for landmark localization that can handle occlusions and deformations better than existing methods. This is essential in scenarios such as bulk aquaculture product monitoring. 
    \item We design a weight regression model that incorporates morphometric features derived from landmarks to improve weight prediction accuracy.
    \item We  perform morphometric shape analyses on five important prawn traits derived from landmark data, and also use Principal Component Analysis (PCA) to find landmarks correlated with shape features.
    \item We evaluate our approach on a large dataset of prawn images and show that it outperforms existing methods in accuracy, robustness, and efficiency.
\end{enumerate}

\section{Method} \label{secmethod}

\begin{figure*}[!t]
\centering
\includegraphics[width=0.98\textwidth]{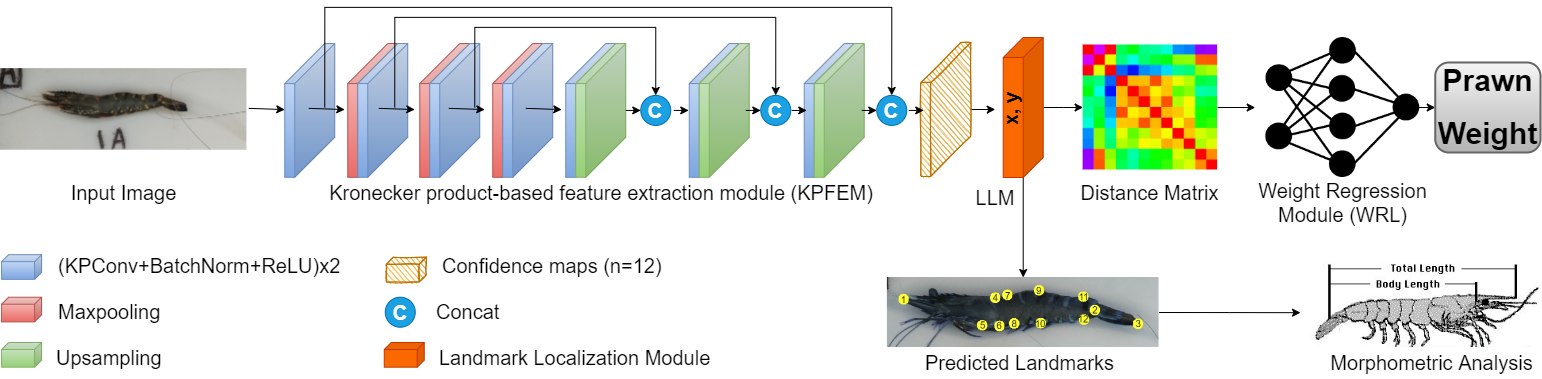}
\caption{An overview of our proposed network architecture used for landmark detection from images that can be used for weight estimation and morphometric analyses. Our architecture consists of two main modules: a Kronecker product-based feature extraction module (KPFEM) and a subsequent landmark localization module (LLM). These are followed by a weight regression module (WRM). The KPFEM module is responsible for extracting features from the input image using Kronecker product-based convolutional layers, while the LLM module localizes the landmarks of the prawn using the extracted features. The WRM module regresses the weight of the prawn based on the detected landmarks. This multi-stage approach has shown promising results for accurate prawn weight and morphometric estimation from images.}
\label{fig_7}
\end{figure*}

Our proposed deep learning architecture is shown in Fig. \ref{fig_7} with two distinct outputs, one for prawn weight, and the second for landmark identification and applied morphometric analyses.

The KPFEM serves as a feature extraction module and utilizes Kronecker convolution operation in a novel network architecture to extract features from the input prawn image. The resulting feature map is then passed to the LLM, which uses a deep learning-based approach to detect 12 landmarks on the prawn body.

The predicted landmarks are then used to calculate the distances between any 12 landmarks, resulting in a total of ($12(12-1)/2 = 66$) possible distances. These distances are then fed to the WRM, which uses a deep learning-based approach to predict the weight of the prawn. The 12 detected landmarks are also used to perform morphometric analysis of the prawn. 
The following subsections provide further details on the role and importance of each of the aforementioned modules in reaching the overall goal of our architecture, i.e. automatic weight estimation and morphometric analysis from prawn images.

\subsection{Kronecker product-based feature extraction module (KPFEM)}
The first module in our architecture is based upon the Kronecker convolution operation implemented within several Kronecker Convolution Layers (KCL), which are explained in detail below. We also provide an explanation of the advantages that KCL provides for our architecture, compared to conventional CNNs. 

\subsubsection{Kronecker Convolution Layer}
One of the main strengths of this work, which makes it suitable for developing a reliable tool for the challenging task of prawn image analysis, is leveraging the Kronecker convolution operation in a unique way to extract more informative features from the input image, resulting in more accurate landmark detection and weight estimation. 

The Kronecker convolution operation is based on the Kronecker product which is a mathematical operation that takes two matrices and produces a new matrix that is formed by multiplying each element of the first matrix with the second matrix.
In the context of convolutional neural networks, the Kronecker product is used to create a weight tensor that is a Kronecker product of two smaller tensors, one that represents the filters in the convolutional layer and another that represents the input image,
to create a large weight tensor. This weight tensor is then used to compute the convolution between the input image and the filter.

In contrast to standard convolutional layers, which apply a single filter across all input channels to find local spatial relationships between input features, 
the Kronecker convolution operation applies a filter that captures both spatial and higher-order information across all input channels. This makes it an effective approach for feature extraction in multi-channel inputs such as RGB images.

The Kronecker product can be written as follows:
\begin{gather}
\label{eq:1}
\begin{array}{c}
\mathop {\left[ {{\bf{A}}_1 } \right]}\limits_{\left( {n \times n} \right)}  \otimes \mathop {\left[ {\begin{array}{*{20}c}
   {}  \\
\end{array}{\bf{F}}_1 \begin{array}{*{20}c}
   {}  \\
\end{array}} \right]}\limits_{\left( {\frac{s}{n} \times \frac{d}{n} \times k \times k} \right)}  
+  \ldots  + \mathop {\left[ {{\bf{A}}_n } \right]}\limits_{\left( {n \times n} \right)}  \otimes \mathop {\left[ {\begin{array}{*{20}c}
   {}  \\
\end{array}{\bf{F}}_n \begin{array}{*{20}c}
   {}  \\
\end{array}} \right]}\limits_{\left( {\frac{s}{n} \times \frac{d}{n} \times k \times k} \right)}  = \mathop {\left[ {\begin{array}{*{20}c}
   {}  \\
   {}  \\
   {}  \\
   {}  \\
   {}  \\
   {}  \\
   {}  \\
\end{array}\begin{array}{*{20}c}
   {}  \\
   {}  \\
   {}  \\
   {}  \\
   {}  \\
   {}  \\
   {}  \\
\end{array}\begin{array}{*{20}c}
   {}  \\
   {}  \\
   {}  \\
   {}  \\
   {}  \\
   {}  \\
   {}  \\
\end{array}{\bf{H}}\begin{array}{*{20}c}
   {}  \\
   {}  \\
   {}  \\
   {}  \\
   {}  \\
   {}  \\
   {}  \\
\end{array}\begin{array}{*{20}c}
   {}  \\
   {}  \\
   {}  \\
   {}  \\
   {}  \\
   {}  \\
   {}  \\
\end{array}\begin{array}{*{20}c}
   {}  \\
   {}  \\
   {}  \\
   {}  \\
   {}  \\
   {}  \\
   {}  \\
\end{array}} \right]}\limits_{\left( {s \times d \times k \times k} \right)}.  \\ 
 \end{array}
\end{gather}
\noindent where matrix $\mathbf{A} \in \bR^{n \times n}$ and the filter matrix $\mathbf{F}$ and the weight matrix $\mathbf{H}$ have the same dimensions: $s$ channels, $d$ filters, and $k \times k$ kernel size.

The Kronecker Convolution Layer (KCL) uses the Kronecker product to arrange convolution filters in a way that reduces the number of parameters by a factor of $1/n$. We explain how this works for different values of $n$ in Eq.~\ref{eq:1}. When $n=1$, we have a real-valued convolution and the Kronecker product is just a scalar multiplication. The filter matrix $\mathbf{F}$ has the same size as the weight matrix $\mathbf{H}$, which is $s \times d \times k \times k$.

When $n=2$, we have a complex-valued convolution and the Kronecker product is between two matrices. The filter matrices $\mathbf{F}_1$ and $\mathbf{F}_2$ are half the size of $\mathbf{H}$, and they contain the filters for each complex component. The algebra is done with matrices $\mathbf{A}_1$ and $\mathbf{A}_2$. This way, we use half as many parameters as in the real case. When $n>2$, we can extend this idea by using smaller filter matrices for each dimension. The size of $\mathbf{H}$ does not change, but the parameter size \texttt{decreases} with higher values of $n$.

The weight tensor $\mathbf{H}$ in the KCL layer is obtained by summing Kronecker products between two groups of learnable matrices. Specifically, it can be expressed as:

\begin{equation}\label{eq:2}
\mathbf{H} = \sum_{i=1}^n \mathbf{A}_i \otimes \mathbf{F}_i,
\end{equation}

\noindent where $\mathbf{A}_i$ is a $n \times n$ matrix that describes the algebra rules, and $\mathbf{F}_i$ is a $ \frac{s}{n} \times \frac{d}{n} \times k \times k$ matrix representing the $i$-th batch of filters. These filters are arranged according to the algebra rules to construct the final weight matrix.

Algorithm \ref{kronecker_conv} is a pseudocode implementation of Kronecker Product Convolution using PyTorch-like syntax. The algorithm takes two input tensors: A, a 2D tensor of shape (height, width), and F, a 4D tensor of shape ($num\_filters$, $num\_channels$, $filter\_height$, $filter\_width$), and performs Kronecker product convolution on them. The output tensor has a shape of ($num\_filters$, $num\_channels$, $output\_height$, $output\_width$), where $output\_height$ and $output\_width$ are calculated based on the size of A and the filter size.

The algorithm works by first computing the Kronecker product of A and F, which is a block matrix of shape ($num\_filters$, $num\_channels$, $heightfilter\_height$, $widthfilter\_width$). This is achieved by expanding the dimensions of A and F and multiplying them element-wise. The resulting block matrix is then reshaped to the desired output shape.

\begin{algorithm}[ht] 
    \label{kronecker_conv}
	\small
	\caption{Kronecker Product, PyTorch-like}
	\LinesNumbered
\begin{lstlisting}[language=Python]
import torch 

def kronecker_product(self, A, F):
    mtx1 = torch.Size("torch.tensor"(A.shape[-2:]) * "torch.tensor"(F.shape[-4:-2]))
    mtx2 = torch.Size("torch.tensor"(F.shape[-2:]))
    res = A.unsqueeze(-1).unsqueeze(-3).unsqueeze(-1).unsqueeze(-1) *
    F.unsqueeze(-4).unsqueeze(-6)
    mtx0 = res.shape[:1]
    out = res.reshape(mtx0 + mtx1 + mtx2)
    return out

\end{lstlisting}
\end{algorithm}

\subsubsection{Standard Convolutional Layer}
A standard convolutional layer convolves the input $\x \in \bR^{t \times s}$ with the filter tensor $\W \in \bR^{s \times d \times k \times k}$ to generate the output $\y \in \bR^{d \times t}$, as follows: 
\begin{equation}
    \y = \text{Conv}(\x) = \W * \x + \mathbf{b},
\label{eq:conv}
\end{equation}
where $s$ is the input channels dimension, $d$ the output, $k$ is the filter size, and $t$ is the input and output dimension. The bias term $\mathbf{b}$ has negligible impact on the number of parameters, resulting in a complexity of $\mathcal{O}(sdk^2)$.

The KCL layer is a convolutional layer that uses a weight tensor $\mathbf{H}$ to organize its filters, which is constructed by summing Kronecker products. The layer can be defined as 
\begin{equation}
    \y = \text{KCL}(\x) = \mathbf{H}*\x + \mathbf{b},
\end{equation}
where $\mathbf{H}$ is a learnable tensor with dimensions $s \times d \times k \times k$. The two groups of learnable matrices used to construct $\mathbf{H}$ are denoted as $\mathbf{A}_n$ and $\mathbf{F}_n$, which are combined through Kronecker products to create $\mathbf{H}$, (see \cref{eq:1} and \cref{eq:2}). The value of $n$ can be set by the user to specify the real or hypercomplex domain, and controls the degree of freedom of $\mathbf{A}_n$ and $\mathbf{F}_n$. The number of parameters in the KCL layer is reduced by a factor of $1/n$ compared to a standard convolutional layer in real-world problems, because typically $sdk^2 \gg n^3$. During training, the matrices $\mathbf{A}_n$ and $\mathbf{F}_n$ are learned and used to construct $\mathbf{H}$. The dimensions of $\mathbf{F}_n$ are $\frac{s}{n} \times \frac{d}{n} \times k \times k$ for squared kernels, and $\frac{s}{n} \times \frac{d}{n} \times k$ for 1D kernels. Hence, The KLC complexity  of the weight matrix can be
approximated to $\mathcal{O}(sdk^2/n)$.

\subsubsection{KCL advantages compared to standard convolution}
Compared to standard convolutional layers, using KCL brings several advantages. As discussed, firstly, the KCL layer reduces the number of parameters by a factor of $1/n$ in real-world problems, where $n$ is the hyperparameter that specifies the desired domain. For example, for RGB images that have $n=3$, the network number of parameters is reduced by $66\%$.
This reduction in parameters can lead to faster training and inference times, as well as reduced memory usage, making KCL  well-suited for resource-constrained devices. Secondly, KCL allows for weight sharing among different channels in multidimensional data, such as colour images, which enables capturing latent intra-channel relations that standard convolutional networks may ignore due to the fixed structure of the weights. This can result in better performance in tasks that involve correlated channels. Finally, the KCL layer can be easily integrated into any convolutional model by replacing standard convolution or transposed convolution operations, and the hyperparameter $n$ provides high flexibility to adapt the layer to any kind of input. Overall, the KCL layer offers a promising alternative to standard convolutional layers and has the potential to improve the performance of convolutional neural networks in various applications.

\subsubsection{Feature Extraction Structure}
The proposed KPFEM module is composed of 14 KCLs that extract features from the input prawn image. As shown in Fig. \ref{fig_7} each layer is followed by a rectified linear unit (ReLU) activation function and every second layer is followed by a max pooling operation to reduce the spatial dimension of the feature maps. The module has "skip connections", which allow information to flow through the network more efficiently by skipping over certain layers that might otherwise impede the flow of information.

\subsection{Landmark Localization Module (LLM)}
The landmark localization module takes the extracted features generated by the KPFEM module and predicts the locations of the landmarks. Therefore, for our keypoint detection task, instead of using \ac{FCN} to directly predict a numerical value of each keypoint coordinate as an output (i.e. regressing images to coordinate values), we modified \ac{FCN} to predict a stack of output heatmaps (i.e. confidence maps), one for each keypoint. The position of each keypoint is indicated by a single, two-dimensional, symmetric Gaussian in each heatmap in the output, and the scalar value of the peak reflects the prediction's confidence score.

Moreover, our proposed LLM not only predicts heatmaps but also predicts scalar values for coordinates of each keypoint. Therefore, during the training process, we have a multi-task loss function, which consists of two losses, i.e. Jensen–Shannon divergence for heatmaps and Euclidean distance for coordinates. The first loss measures the distances between the predicted heatmaps and the ground-truth heatmaps, while the second loss measures the distances between the predicted coordinates and the ground-truth coordinates. Then, we take the average of the two losses as the optimization loss.

As demonstrated in Fig. \ref{fig_7}, the LLM output is a set of predicted landmark coordinates, using which also a distance matrix is produced to feed to the next module in our proposed architecture. 

\subsection{Weight Regression Module (WRM)}
The final component of our architecture is the weight regression module which is made up of a multilayer perceptron (MLP) that consists of five layers of nodes: an input layer, three hidden layers, and an output layer. Each node in the hidden layers applies the ReLU activation function to the weighted sum of inputs from the previous layer.

The weight regression module takes the output of the landmark localization module, which includes the predicted locations of the landmarks, and measures 66 distances between any 12 landmarks to predict the weight. We use the distances between the landmarks to estimate significant traits such as total length, body length, carapace length, and length-width ratios. These measurements are typically made between easily distinguishable landmarks. The distances between them are assigned a trait name and then used as inputs for a pre-trained regression model that maps these estimates to the prawn weight. Specifically, we generated morphological measurements using a combination of methods from previous studies and those that were possible within the constraints of the available images. For the morphometric shape analysis, we used 8,164 photographed specimens to estimate 66 morphometric distances derived from 12 landmarks. These distances were then used by the weight regression module to predict the prawn weight directly from the landmarks. This is a practical and useful application of our model. We trained our model using a mean squared error (MSE) loss to minimize the difference between the predicted weight and the ground-truth weight. This ensures that our model is accurate and reliable for predicting prawn weights based on landmark information.

\begin{figure*}[ht]
\centering
\includegraphics[width=0.78\textwidth]{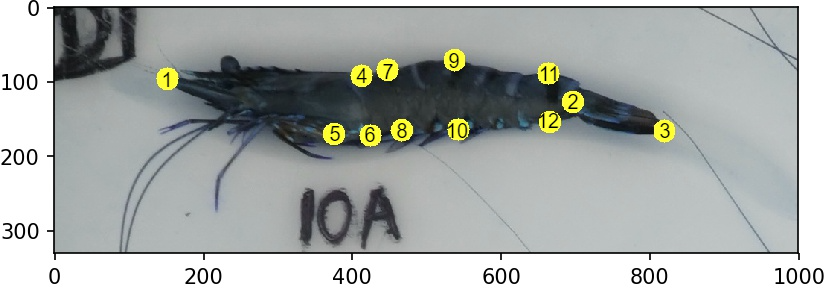}
\caption{Keypoints (landmarks) of interest marked on the body of the Black Tiger Prawn (\textit{Penaeus monodon})}
\label{fig_5}
\end{figure*}

\section{Experiments} \label{secresult}
In this section, we present the experimental setup and results of our proposed approach for automated morphometric analysis and weight estimation of prawns from images. We first describe the dataset used for the experiments and the data preprocessing steps. Next, we provide details on the training and evaluation of our approach and compare it to existing methods. Finally, we analyze the results and discuss the strengths and limitations of our approach.

\subsection{The Dataset} \label{secdata}
We collected images of 8164 individual Black Tiger Prawns (\textit{Penaeus monodon}) from an aquaculture farm in Australia. The images were captured using a digital camera under natural lighting and saved as JPG files. We used ImageJ software to annotate each image with 12 landmarks that correspond to prawn size and shape (\cref{fig_5}). The landmark positions were defined based on a previous study \cite{hung2014} and adapted to our image quality.

To evaluate the reliability of the landmark data, we measured 300 images twice by one operator (O1) and once by another operator (O2) and computed the intraclass correlation coefficient (ICC) for each landmark pair using R software. The ICC values ranged from 0.96 to 0.99, indicating high agreement between operators. The annotated images contain 12 specific and homologous points on the prawn body, including eyes, antennae, and tail. For a complete list of the 12 landmarks,  see \cref{table1}. We then derived 5 traits from these landmarks, which are listed in  \cref{table2} and used for morphometric analyses in \cref{secmorph}.

The dataset, therefore, includes 
for each image, the 12 ground-truth landmark coordinates, and morphometric measurements obtained from the annotated coordinates on images. 
The dataset was divided into three parts: training, validation, and testing. We used the training dataset to train the Deep Learning models, the validation dataset to assess model performance during training, and the testing dataset to evaluate the performance of the trained models. The ground truth landmarks for each prawn image had the form $[(x_1,y_1), .. ,(x_k,y_k)]$, where $(x_i,y_i)$ represented the $ith$ landmark location. Our model was then trained to predict keypoint locations for each prawn image as shown in  \cref{fig_5}. The predicted landmark had the same form as the ground truth. 

Using a well-curated and annotated dataset is critical for the success of Deep Learning-based morphometric analysis. It provides the models with sufficient training data to learn shape information from the images and accurately identify landmark points. To achieve this, 
we carefully curated the dataset to ensure a balanced representation of different prawn weights and body shapes, to enhance the robustness of the results. We also pre-processed the images to make sure they were of similar size and resolution and to eliminate background noise and other distractions that could impact the accuracy of landmark identification.

\begin{table*}[!t]
\centering
\caption{Lists of the 12 landmarks used in our analysis and their descriptions}
\label{table1}
\small
\begin{spacing}{1.29}
\setlength{\tabcolsep}{15pt}
\begin{tabular}{ll}
\hline
\textbf{Landmark number} & \textbf{Landmark Description} \\
\hline
1 & Placed on the most anterior point of the antennal scale \\
2 & Placed on the most anterior point of the tail \\
3 & Placed on the most posterior point of the tail \\
4 & Placed on the junction of the carapace and abdomen placed at the most dorsal point \\
5 & Placed on the midway along the carapace on the ventral side of the prawn \\
6 & Placed on the junction of the carapace and abdomen placed at the most ventral point \\
7 & Placed dorsally on the midpoint of the first abdominal segment \\
8 & Placed ventrally on the midpoint of the first abdominal segment \\
9 & Placed dorsally on the midpoint of the third abdominal segment \\
10 & Placed ventrally on the midpoint of the third abdominal segment \\
11 & Placed dorsally on the midpoint of the last abdominal segment \\
12 & Placed ventrally on the midpoint of the last abdominal segment \\
\hline
\end{tabular}
\end{spacing}
\vspace{-1em}
\end{table*}

\subsection{Evaluation Metrics}\label{secmetrc}
In this section, we describe the performance metrics used to optimize and evaluate the model and compare the quality of the predicted keypoint locations.

\subsubsection{Euclidean distance}
The first metric is the Euclidean distance, which measures the distance of the keypoints based on their coordinates and does not depend on how the ground truth has been determined. A value of 0 indicates that the predicted keypoint is exactly at the same coordinate as the ground truth keypoint. We calculate the sum of the squared Euclidean distance of the difference between two kyepoints
i.e., the predicted keypoint and the ground truth human-annotated keypoint 
This represents the total difference between the two points. The equation for Euclidean distance is shown in Equation \ref{eq:Euclidean}, 
\begin{equation}
d(g,p)= \sqrt{\sum_{i=1}^n  ( v_{i}^{g} - v_{i}^{p} )^{2}},
\label{eq:Euclidean}
\end{equation}
where $g$ and $p$ are two sets of points in Euclidean n-space for ground truth and prediction, respectively, $ v_{i}^{g} , v_{i}^{p}$ are Euclidean vectors starting from the origin of the space (initial point) for the ground truth and prediction, respectively, and $n$ is the number of keypoints.

\subsubsection{Jensen-Shannon divergence}
The second metric is the Jensen-Shannon divergence, which is a distance measure between two distributions and can be used to quantify the accuracy of the predicted keypoints distribution compared to the ground-truth keypoints distribution. The lower the value, the better the model performs. This distance is calculated based on the Kullback-Leibler divergence (KLD) and can be expressed as shown in Equation \ref{eq:kld}.
\begin{equation}
KLD(P||Q) =  \sum_{i=1}^n p_i(x)log \left(\frac{p_i(x)}{q_i(x)}\right),
\label{eq:kld}
\end{equation}
where The $KLD$ for two probability distributions $P$ and $Q$, and when there are $n$ pairs of predicted $p$, and ground truth $q$.

The Jensen-Shannon divergence (JSD) is then calculated using Equation \ref{eq:Jensen}, 
\begin{equation}
JSD_M(P||Q) = \sqrt{\frac{KLD(p \parallel m) + KLD(q \parallel m)}{2}},
\label{eq:Jensen}
\end{equation}
where $m$ is the point-wise mean of $p$ and $q$. The JSD measures the difference between two probability distributions, with a value of 0 indicating no difference between the distributions.

\begin{table*}[!t]
\centering
\caption{Important prawn traits, their descriptions, and their corresponding landmark coordinates from \cref {table1}}
\label{table2}
\small
\begin{spacing}{1.29}
\setlength{\tabcolsep}{10pt}
\begin{tabular}{l c >{\raggedright\arraybackslash}p{7cm}}
\hline
\textbf{Trait} & \textbf{Landmark Coordinates} & \textbf{Trait Description} \\
\hline
Total length & 1-3 & From the most anterior point of the antennal scale to the most posterior point of the tail \\
Body length & 1-2 & From the most anterior point of the antennal scale to the most anterior point of the tail \\
First abdominal segment height & 7-8 & From the dorsal midpoint to the ventral midpoint on the first abdominal segment \\
Third abdominal segment height & 9-10 & From the dorsal midpoint to the ventral midpoint on the third abdominal segment \\
Last abdominal segment height & 11-12 & From the dorsal midpoint to the ventral midpoint on the last abdominal segment \\
\hline
\end{tabular}
\end{spacing}
\vspace{-1em}
\end{table*}

\subsubsection{Object Keypoint Similarity (OKS)}
The third metric use to evaluate the performance of our landmark detection model is the Object Keypoint Similarity (OKS), which is determined by dividing the distance between expected and ground truth points by the object's scale. OKS keypoints estimation serves the same purpose as Intersection over Union ($IoU$) as in object detection. This gives the similarity between the keypoints of the two detected boxes, with a result between 0 and 1, where 0 means no similarity between the keypoints, while perfect predictions will have OKS=1. The equation for OKS is shown in Equation \ref{eq:oks}. 
\begin{equation}
OKS = \frac{\sum_{i} \exp \left(-d_{i}^{2} / 2 s^{2} k_{i}^{2}\right) \delta\left(v_{i}>0\right)}{\sum_{i} \delta\left(v_{i}>0\right)},
\label{eq:oks}
\end{equation}
where $d_{i}$ is the Euclidean distance between the detected keypoint and the corresponding ground truth, $v_{i}$ is the visibility flag of the ground truth, $s$ is the object scale, while $k_{i}$ represents a per-keypoint constant that controls falloff.

Equations \ref{eq:Euclidean} and \ref{eq:Jensen} were used for model training and optimization and also used to compare different models' performance, as shown in Table \ref{tab:Params}. Equation \ref{eq:oks} was used as a final evaluation metric for all the models used in this study, as shown in Table \ref{tab:results}.

\subsubsection{Precision and Recall}

Object Keypoint Similarity (OKS) \citep{Lin2014b} was used as a performance metric (see section \ref{secmetrc} for more details).
As explained, the following 6 metrics are usually used for characterising the performance of a keypoint detector model and were applied in our study.
\begin{itemize}
    \item Average Precision ($AP$):
        \begin{itemize}
            \item $AP$ ~~ ( at $OKS=.50:.05:.95$ (primary metric))
            \item $AP^{.50}$ ( at $OKS=.50$ )
            \item $AP^{.75}$ ( at $OKS=.75$ )
        \end{itemize}
    \item Average Recall ($AR$):
        \begin{itemize}
            \item $AR$ ~~ ( at $OKS=.50:.05:.95$)
            \item $AR^{.50}$ ( at $OKS=.50$)
            \item $AR^{.75}$  ( at $OKS=.75$)
        \end{itemize}
\end{itemize}

\subsection{Model training} \label{sectrain}
In this section, we will describe the process of training our Deep Learning model to identify landmarks in prawn images. Our model architecture, which was described in 
Section \ref{secmethod}, was trained using the aforementioned large dataset of annotated prawn images. The training process consisted of several key steps, including data preprocessing, model selection, and hyperparameter tuning.

\subsubsection{Data Preprocessing}
Before the training process can begin, the image data must be preprocessed to ensure that it is suitable for input into the Deep Learning model. This includes resizing the original image size of $1000\times331$, to a consistent size of $320\times320$, normalizing the pixel values, and converting the images to grayscale if necessary. In addition, the annotated landmark locations must be converted into a format that can be used by the model, such as a heatmap or a set of points.
We applied some image transformation operations to our training set with certain probabilities. These operations are: Horizontal flip: $0.5$, Vertical flip: $0.5$, Shift and scale: $0.5$ (shift limit = $0.0625^{\circ}$, scale limit = $0.20^{\circ}$), Rotation: $0.5$ (rotation limit = $20^{\circ}$), Blur: $0.3$ (blur limit = $1$), RGB-shift: $0.3$ (R-shift limit = $25$, G-shift limit = $25$, B-shift limit = $25$).
These operations help to improve the robustness of our model to lighting changes.
We did not transform the images in our validation or test sets in any way.

The ground truth landmarks for each prawn image are represented in the form of $[(x_1,y_1), .. ,(x_k,y_k)]$, where $(x_i,y_i)$ denotes the location of the $ith$ landmark. In the training process, both the original $(x_i,y_i)$ values and the converted heatmap derived from these values are utilized in the loss function.

\subsubsection{Model Selection}
Once the data has been preprocessed, the next step is to select a model architecture that is suitable for our problem. There are a variety of Deep Learning models that can be used for image landmark identification. In this work, we have selected   six models for our experiments: U-net \citep{Ronneberger2015}, ResNet-18 \citep{He2015ResNet}, ShuffleNet-v2 \citep{Zhang2018b}, MobileNet-v2 \citep{MobileNetV2}, SqueezeNet \citep{Iandola2016}, and our proposed KPFEM.

\subsubsection{Hyperparameter Tuning}
Once the model architecture has been selected, the next step is to tune the hyperparameters to achieve optimal performance. This includes selecting the optimal batch size, learning rate, and number of epochs. The hyperparameters are selected using a combination of grid search and cross-validation to ensure that the model is generalizing well and not overfitting to the training data.
For this problem set, we chose a learning rate of $1 \times 10^{-3}$ as the best option. All models took about $200$ epochs to train on this problem and we reduced the learning rate by $\gamma = 0.001$ after every 50 epochs.  We also used Adam optimiser \cite{Kingma2014Adam:Optimization} with $\beta_1 = 0.9$, $\beta_2 = 0.999$, and $\epsilon = 1.0 \times 10^{-08}$. We applied these hyperparameters to all six models. The best model configuration may vary depending on the application, so these results do not cover all possible model configurations.

We split the dataset into three sets: "Train", "Validation", and "Test", comprising 40\%, 20\%, and 40\% of the data, respectively.
We trained all models on the Train subset of the data with the same hyperparameters. All models had two outputs (heatmap and coordinates) with two losses (see \cref{secmetrc}). All models took $320\times320$ input images and produced $56\times56$ output heatmaps except U-net which had $320\times320$ output images.

\subsubsection{Training}
The final step in the training process is to train the model using the preprocessed data and the optimized hyperparameters. The model is trained using a supervised learning approach, where the ground truth landmark locations are used to calculate the loss function and update the model parameters. The training process is repeated until the model has reached convergence or a maximum number of epochs has been reached. We used Pytorch framework \cite{Paszke2019} on a Linux host with a single NVidia GeForce RTX 2080 Ti GPU and a batch size of $64$.

Once the model was trained, it was evaluated on the test subset of the dataset to assess its performance in identifying landmarks in new, unseen images. The evaluation metrics, described in the evaluation metrics section (\cref{secmetrc}), are used to quantify the accuracy of the model and provide insight into its strengths and weaknesses.

\section{Results} \label{secresy}
In this section, we present the results of our experiments on prawn  landmark detection, used for weight estimation and morphometric analyses from images using our proposed approach. 
We conducted experiments to optimize our method and assess its performance against the five other models that we mentioned earlier, based on image throughput (speed), accuracy, inference time, and generalization ability. The test subset was used to measure these models (see section \ref{secdata} for more information).

\subsection{Landmark Detection}
Table \ref{tab:Params} presents a comparative analysis based on several metrics, including the number of floating-point operations (FLOPs), the number of parameters, model size, and throughput in images per second, as well as the coordinates loss (Equ. \ref{eq:Euclidean}), heatmap loss (Equ. \ref{eq:Jensen}), and the average of both losses. 
It is worth noting that, the actual throughput of a network is the number of instances it can process in one second with the optimal batch size. It is obtained by dividing the total number of instances processed by the total time taken to process them and can vary depending on various factors such as the model's complexity, the input data size, and the available hardware resources. 

 \begin{table}[htbp]
\centering
\caption{Landmark Detection Performance Comparison of our model compared to five benchmark models.}
\resizebox{\linewidth}{!}{%
\begin{tabular}{l|cccc|ccc}  
\toprule
Network              & \shortstack{FLOPs \\ (x$10^6$)}         & \shortstack{\#Params \\ (x$10^6$)}   & \shortstack{Size \\ (MB)}  & \shortstack{Throughput \\ (img/sec)} & Coords & HeatMap & Avg. \\ 
\midrule
U-net \citep{Ronneberger2015}& 16.52 & 31.04 &   124.3     &    201     &  0.024 &  0.355   & 0.190    \\ 
ResNet-18 \citep{He2015ResNet}& 2.62 & 12.85 &   51.5    &     404  &  0.028 &   0.090   &  0.059   \\  
ShuffleNet-v2 \citep{Zhang2018b}& 0.44 & 3.06 &   12.5     &    170     &  0.047 &   0.153  &  0.100   \\  
MobileNet-v2 \citep{MobileNetV2}& 0.67 & 4.10 &   16.7    &    205    &  0.041 &   0.137   &  0.089   \\  
SqueezeNet \citep{Iandola2016}& 0.92 & 2.33 &   9.4    &    551    &  0.027 &   0.078   &  0.052   \\  
\midrule
{KPFEM (ours)} & \textbf{0.01}  & \textbf{0.39} &   \textbf{1.6}     &    \textbf{562}      &  0.023 &   0.084   &  0.0053   \\  
\bottomrule
\end{tabular}
}
\label{tab:Params}
\end{table}

The experimental results shown in \cref{tab:Params}, were conducted on a desktop computer equipped with a single NVidia GeForce RTX 2080 Ti GPU. These results demonstrate that our proposed network which is based on our KPFEM feature extraction method outperforms other networks in several aspects, including having the lowest number of parameters (47 times fewer parameters than U-net \citep{Ronneberger2015}), the smallest size on the hard disk, and the second-highest throughput, following SqueezeNet \citep{Iandola2016}. Moreover, our model exhibits a lower average loss than other popular models, including U-net \citep{Ronneberger2015}, ShuffleNet-v2 \citep{Zhang2018b}, and MobileNet-v2 \citep{MobileNetV2}.

The low number of parameters, small model size, and high throughput of our model make it a promising solution for many real-time applications, such as mobile prawn video processing and portable phenotyping systems \citep{saleh2022a}. However, there are some limitations that must be taken into consideration when interpreting the results. First, our experiments were conducted on a single GPU, which may not represent the complete profile performance, and may vary on other hardware configurations. Second, the dataset used for the experiments was limited in size and scope, and therefore, further studies on more diverse datasets are necessary to validate the effectiveness and generalizability of our approach. Future work will focus on addressing these limitations and further improving the efficiency and accuracy of our approach.

Table \ref{tab:results} shows the performance of our model on the test subset of the dataset compared to benchmark models in landmark detections, using the OKS evaluation metric. 
To assess the generalization effectiveness of our model, 
we compared the performance of our model with randomly initialized weights, against the other models with randomly initialized weights as well to provide a direct comparison.
Table \ref{tab:results} demonstrates that our proposed model performs well in generalization with only 40\% of the data and without the use of transfer learning, when combined with robust data augmentation techniques.

The overall outcome depicted in Table \ref{tab:results} indicates that our proposed network surpasses both ShuffleNet-v2 \citep{Zhang2018b} and MobileNet-v2 \citep{MobileNetV2} in terms of landmark detection performance, with an accuracy of $AP=0.986$, while still competing with SqueezeNet \citep{Iandola2016} even though it has substantially fewer parameters. It is noteworthy that our model attains this high accuracy with only 0.39M parameters and without relying on transfer learning. These results clearly demonstrate the effectiveness and generalisability of our KPFEM-based model.

\begin{table}[htbp]
\centering
\caption{Performance comparison using the $OKS$ metric on the \textbf{test} dataset.}
\resizebox{\linewidth}{!}{%
\begin{tabular}{l|ccc|ccc}
\toprule
Network                      & $AP$ & $AP^{.50}$ & $AP^{.75}$ & $AR$ & $AR^{.50}$ & $AR^{.75}$   \\ 
\midrule
U-net \citep{Ronneberger2015}  & 0.981 &   0.990   &  0.990      &  0.974 &  0.999    & 0.999    \\ 
ResNet-18 \citep{He2015ResNet} & 0.984 &   0.990   &   0.990     &  0.982 &   0.999   &  0.999   \\  
ShuffleNet-v2 \citep{Zhang2018b}& 0.957 &   0.990   &   0.990     &  0.979 &   0.999   &  0.999   \\  
MobileNet-v2 \citep{MobileNetV2}& 0.963 &   0.990   &   0.989     &  0.979 &   0.999   &  0.996   \\  
SqueezeNet \citep{Iandola2016} & 0.971 &   0.990   &   0.990     &  0.983 &   0.999   &  0.999   \\  
\midrule
{KPFEM (ours)}      & 0.986 &   0.990   &   0.990     &  0.985 &   0.999   &  0.999   \\  
\bottomrule
\end{tabular}
}
\label{tab:results}
\end{table}

\subsection{Weight Estimation}
\begin{figure*}[!t]
\centering
\includegraphics[width=0.98\textwidth]{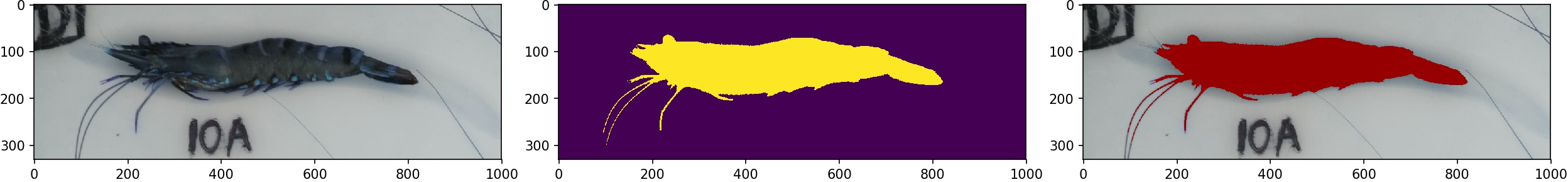}
\caption{Example of threshold segmentation. From left to right: (a) original image, (b) segmented prawn body, (c) overlay of the segmentation on the original image.}
\label{fig_4}
\end{figure*}

We compared our proposed approach with two existing methods for prawn weight estimation: the traditional linear regression method \cite{Konovalov2019b}, and a Deep Learning-based method \cite{Bravata2020} based on shape or contour features for weight estimation.
In the shape or contour features method for prawn weight estimation, the prawn body is segmented from the image using a threshold segmentation process. This process involves selecting a threshold value that separates the pixels in the image into two groups: foreground and background. The pixels with intensity values above the threshold are classified as foreground pixels, which belong to the prawn body, while the pixels with intensity values below the threshold are classified as background pixels. This is demonstrated in \cref{fig_4}.

After segmenting the prawn body, the next step is to estimate its weight. One way to do this is to count the number of pixels in the segmented region, which is assumed to be proportional to the prawn weight. This pixel count is then correlated to the actual weight of the prawn using a linear regression model ( e.g using a mathematical model similar to \cite{Konovalov2019b} for fish weight estimation). Another approach is to use Deep Learning ( e.g. a neural network \cite{Bravata2020} that predicts the prawn's weight based on the pixel count).
However, these methods have some limitations. For example, it assumes that the relationship between pixel count and prawn weight is linear. Additionally, this method does not take into account the shape and position of the prawn body, which can vary from one image to another.

\begin{figure*}[ht]
\centering
\includegraphics[width=0.68\textwidth]{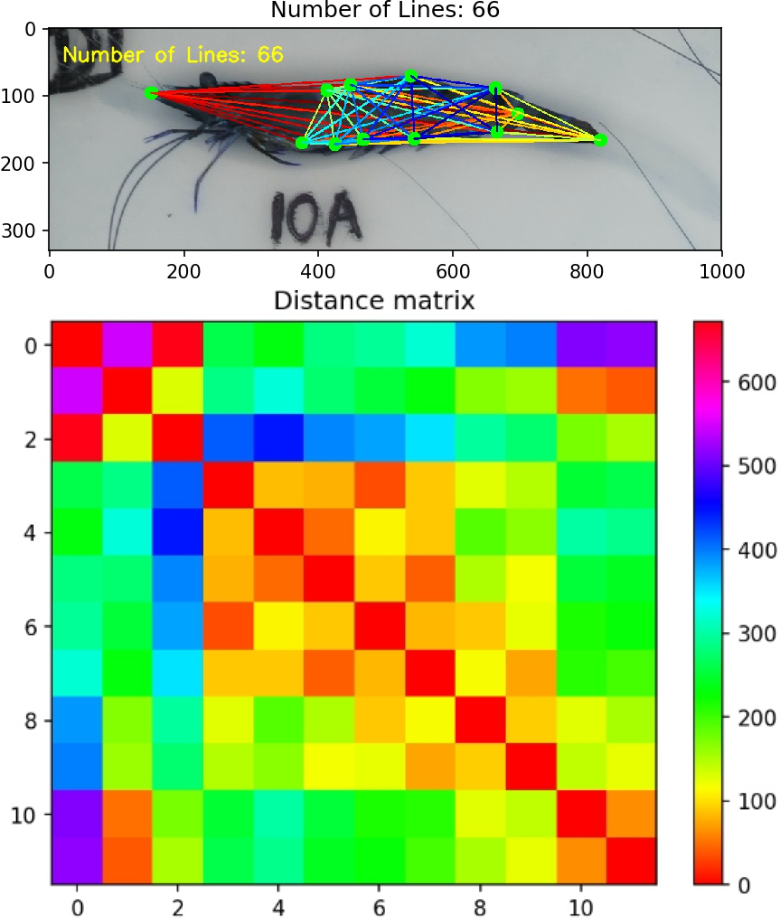}
\caption{Top: Image of a prawn with the 66 possible distances between its 12 landmarks marked. Bottom: The resulting distance matrix plot computed from these distances.}
\label{fig_6}
\end{figure*}

These methods were compared against our proposed approach for prawn weight estimation, which involves generating 66 possible distances from the 12 landmarks and feeding them into a weight regression module. 
We generated a distance matrix as seen in \cref{fig_6} using the 12 predicted landmarks. This distance matrix contains the pairwise Euclidean distances between all possible pairs of the 12 landmarks, resulting in a total of ($12(12-1)/2 = 66$) possible distances. These distances are then fed to the WRM, which uses a deep learning-based approach to predict the weight of the prawn. Our proposed approach of using the distance matrix as features for prawn weight estimation has several advantages over traditional methods such as the  segmentation method. It takes into account the spatial relationships between the landmarks, providing a more accurate estimation of prawn weight. Furthermore, our approach is less sensitive to variations in the prawn's posture or orientation, which can significantly affect the accuracy of traditional methods.

Table \ref{table:comparison} shows the comparison results. Our proposed approach achieved the lowest mean absolute error (MAE) and mean squared error (MSE) values and the highest coefficient of determination among the three methods. Specifically, our approach achieved an MAE of 0.649 g, an MSE of 0.986 g, and a coefficient of determination of 0.934, which outperformed the other two methods. These results demonstrate the effectiveness and superiority of our proposed approach for prawn weight estimation. 

\begin{table}[htbp]
\centering
\caption{Comparison of prawn weight estimation methods using mean absolute error (MAE) in grams, mean squared error (MSE) in grams, and coefficient of determination ($R^2$)}
\begin{tabular}{lccc}
\hline
\textbf{Method} & \textbf{MAE (g)} & \textbf{MSE (g)} & \textbf{Coefficient of determination}\\
\hline
Linear Regression & 0.893 & 1.848 & 0.825\\
Deep Learning-based Method & 0.880 & 1.713 & 0.896\\
Proposed Approach & \textbf{0.630} & \textbf{0.735} & \textbf{0.952}\\
\hline
\end{tabular}
\label{table:comparison}
\end{table}

\begin{figure*}[ht]
\centering
\includegraphics[width=0.98\textwidth]{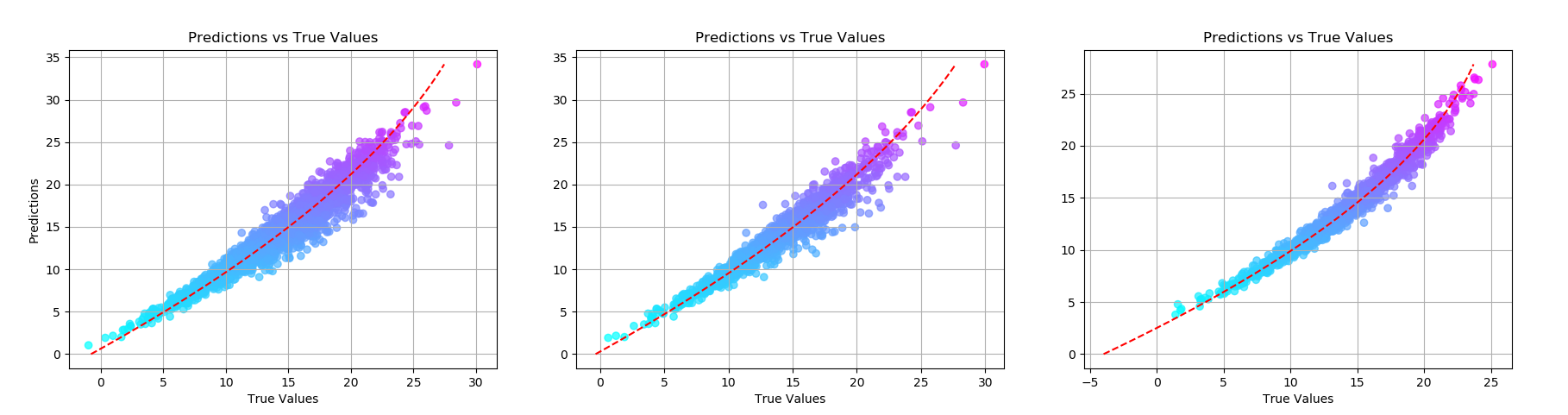}
\caption{Plot showing the relationship between predicted weight and true weight for three different methods. From left to right: results for the Linear Regression method, Deep Learning-based Method, and our Proposed Approach. }
\label{fig_1}
\end{figure*}

In addition to evaluating the overall performance of our proposed weight estimation approach, we also used visualizations to further understand our model's performance and identify areas for improvement. Specifically, we used \cref{fig_1} to visualize the relationships between the predicted and the true weight values. 
The plot in \cref{fig_1} shows the relationship between predicted weight and true weight for three different methods: Linear Regression, Deep Learning-based Method, and the Proposed Approach. The plot can provide information about the accuracy and precision of the different methods. For example, if the points on the plot fall close to the diagonal line ($y = x$), then the predicted weights are close to the true weights, indicating a high level of accuracy. Additionally, if the points are tightly clustered around the diagonal line, then the method is precise in its predictions.

Based on the plots in \cref{fig_1}, it appears that the Proposed Approach has a higher correlation between predicted and true weight values compared to the Linear Regression and Deep Learning-based methods. This can be seen by the tighter clustering of points around the line of best fit. The presence of outliers in the Linear Regression and Deep Learning-based methods may be affecting the overall correlation. 

Outliers can affect correlation in a number of ways. They can either increase or decrease the correlation coefficient depending on their location relative to the regression line. An outlier that is near where the regression line might normally go increases the correlation value, while an outlier away from the regression line decreases it.

\subsection{Ablation Study on Weight Estimation} \label{secabs}
Our proposed model uses 66 distances as its features to predict the prawn weights. To investigate the impact of dimensionality reduction on our proposed approach, we conducted an ablation study using Principal Component Analysis (PCA). The purpose of PCA was to reduce the high number of dimensions/features per observation, which in our case were the 66 possible distances between the 12 landmarks marked on a prawn.
We applied PCA to the distance matrix and reduced the number of components to 2, 5, and 10. We then used the reduced feature sets to predict 
the weight of prawns. The results are shown in \cref{table:pca}.

Surprisingly, we found that using the full set of 66 distances without PCA resulted in better weight estimation results. This suggests that PCA was not necessary for this particular task and that the high-dimensional feature space was important for accurately capturing the information needed for prawn weight estimation. This finding is also in agreement with the observation shown in \cref{table:pca}, which demonstrates the more features used, the lower the MAE and MSE of the model.

 \begin{table}[htbp]
\centering
\caption{Comparison of weight estimation results using PCA with a different number of components.}
\begin{tabular}{lccc}
\hline
\textbf{Method} & \textbf{MAE (g)} & \textbf{MSE (g)} & \textbf{Coefficient of determination}\\
\hline
PCA ($n=2$)   & 0.784 & 0.851 & 0.915\\
PCA ($n=5$)   & 0.765 & 0.842 & 0.924\\
PCA ($n=10$)  & 0.724 & 0.816 & 0.931\\
No PCA & \textbf{0.630} & \textbf{0.735} & \textbf{0.952}\\
\hline
\end{tabular}
\label{table:pca}
\end{table}


\subsection{Morphometric Analyses}  \label{secmorph}
In our proposed 
deep learning model, we detect keypoints (landmarks) on the prawn body to achieve highly accurate weight estimations. By detecting and analysing these landmarks, other important physical characteristics of the prawn, such as its length, width, and shape, can be extracted. The best part is that this morphometric analysis is essentially a byproduct of the weight estimation process. This represents a significant value-add for aquaculture and fisheries managers, who can now obtain valuable morphometric information about their prawn populations without incurring additional expenses or resources.

Morphometric analyses refer to the measurement of various physical features of an organism, such as length, width, and area. These measurements are essential for understanding the biology of the organism and can be used to identify and classify different species while understanding their growth rate and possible body deformities.
Morphometric information can also be used to study and identify specific traits that are desirable for selective breeding or commercial purposes. 

Traditionally, morphometric analyses involve having a human operator measure and manually record various morphology traits on the animal body. This is a slow and inefficient process that cannot be easily scaled.
Automated morphometric analyses facilitated using deep-learning-based computer vision and image processing, on the other hand, offer several benefits over traditional manual measurements. Firstly, they are much faster and more efficient, allowing researchers and operators to process larger stock/data quantities in a shorter amount of time. This is particularly important when dealing with large populations/datasets, as manual measurements are time-consuming and error-prone. Additionally, automated morphometric analyses are more objective, as they are not subject to human bias or error.

\begin{figure*}[htbp]
\centering
\includegraphics[width=0.98\textwidth]{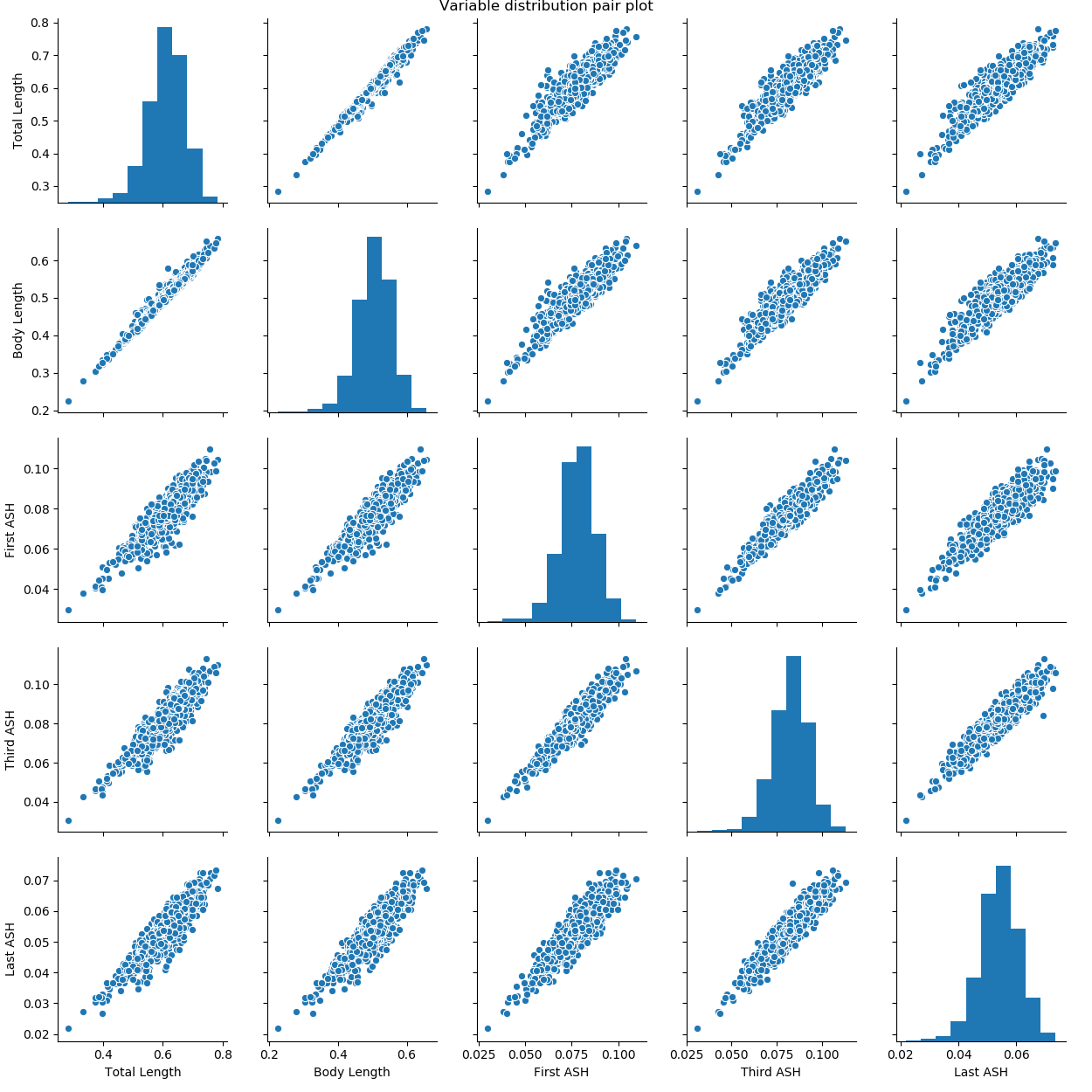}
\caption{The distribution pair plots for  the five important prawn traits from \cref {table2}, i.e. Total length, Body length, First abdominal segment height "First ASH", Third abdominal segment height "Third ASH", Last abdominal segment height "Last ASH"}
\label{fig_8}
\end{figure*}

\begin{figure*}[htbp]
\centering
\includegraphics[width=0.98\textwidth]{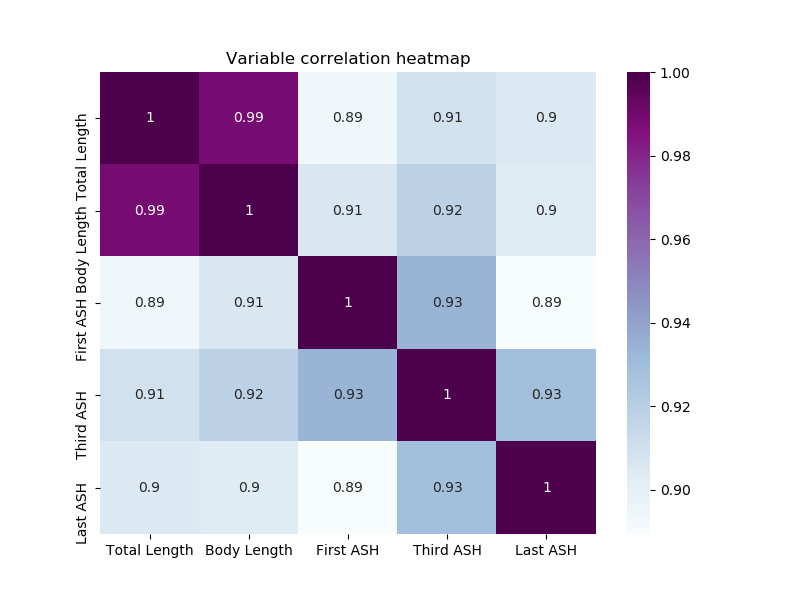}
\caption{The correlation heatmap for  the five important prawn traits from \cref {table2}, i.e. Total length, Body length, First abdominal segment height "First ASH", Third abdominal segment height "Third ASH", Last abdominal segment height "Last ASH"}
\label{fig_10}
\end{figure*}

We performed a morphometric shape analysis on five important prawn traits derived from landmark data. These traits are shown in \cref{table2} and include total length, body length,  the first abdominal segment height (First ASH),   the third abdominal segment height(Third ASH), and   the last abdominal segment height (Last ASH).
We used the landmark data obtained from our model to calculate these five traits for each individual prawn and plotted their distributions in \cref{fig_8}. We also computed the correlation matrix among the traits and visualized it as a heatmap in \cref{fig_10}. The results show that total length and body length are highly correlated ($r = 0.99$), and so are the First and Third, and Third and Last ASH ($r = 0.93$). The other traits had high correlations as well. These patterns reflect the variation in shape and size among the prawns.

The accuracy of measurements is critical for the quantitative comparison of prawn morphometry, where accuracy refers to the closeness of measurements to the true value. To assess accuracy, this study utilized the mean absolute difference (MAD) between manual and DL measurements.

The MAD is a measure of accuracy that calculates the average absolute deviation between two values. It is obtained by dividing the sum of absolute deviations between each value by the number of values. The mathematical expression for MAD is:
$$MAD = \frac{1}{n}\sum_{i=1}^{n}\left|(x_i-y_i)\right|$$ where $n$ is the total number of observations, $x_i$ and $y_i$ are the values of the $i-th$ observation in two different samples, and the vertical bars denote absolute value.

Table \ref{tab:morph} presents a performance comparison of various deep learning (DL) networks in terms of their ability to accurately measure morphometric features. The table shows the mean absolute difference (MAD) in millimetres between manual and DL measurements for total length, body length, first ASH, third ASH, and last ASH. The results indicate that our proposed KPFEM-based network outperforms other networks such as U-net, ResNet-18, ShuffleNet-v2, MobileNet-v2, and SqueezeNet in terms of MAD for all morphometric features. This suggests that our approach is more accurate in measuring morphometric features compared to other DL networks used for landmark detection.

\begin{table}[htbp]
\centering
\caption{Morphometric Analyses Performance Comparison: Mean Absolute Difference (MAD) between Manual and DL Measurements (mm)}
\resizebox{\linewidth}{!}{%
\begin{tabular}{l|ccccc}
\toprule
Network  & Total length & Body length & First ASH & Third ASH & Last ASH    \\ 
\midrule
U-net \citep{Ronneberger2015}     & 1.81 & 1.83 & 1.73 & 1.72 & 1.51  \\ 
ResNet-18 \citep{He2015ResNet}    & 1.72 & 1.85 & 1.71 & 1.64 & 1.43  \\  
ShuffleNet-v2 \citep{Zhang2018b}  & 2.34 & 2.22 & 2.15 & 2.15 & 2.02  \\  
MobileNet-v2 \citep{MobileNetV2}  & 2.23 & 2.24 & 2.04 & 2.12 & 1.94  \\   
SqueezeNet \citep{Iandola2016}    & 2.14 & 1.93 & 1.94 & 1.85 & 1.74  \\   
\midrule
{KPFEM (ours)}  & \textbf{1.61} & \textbf{1.54} & \textbf{1.53} & \textbf{1.45} & \textbf{1.31}  \\    
\bottomrule
\end{tabular}
}
\label{tab:morph}
\end{table}

\subsection{Principal Component Analysis (PCA)}
In addition to our aforementioned analysis, we also performed PCA on the predicted landmarks from our model to analyze the morphometric shape of prawns. PCA is a statistical technique commonly used for dimensionality reduction and identifying patterns in data. By reducing the dimensionality of the data, PCA allows us to identify the principal components (PCs) that explain the most variability in the data.

We chose to use PCA in our analysis as a data-driven approach to analyze prawn shape variation in the predicted landmark data. Even though we have ground truth measurements of body size for the prawns, 
PCA allows us to understand the PCs and explain variability in the data and how it represents the major sources of shape variation in  prawns. This provides a complementary and holistic view of the shape variation beyond individual distance measurements.

It is important to note that the results obtained from PCA are based on the predicted landmark data, which may have some inaccuracies. However, we believe that the predicted landmark data still provide valuable insights into the shape variation among prawns and can be used as a proxy for understanding the morphometric characteristics of the prawns. This is especially important because collecting manual landmarks is impractical at large scales. 

The PCA analysis revealed that the first two principal components (PC1, PC2) accounted for 94.2\% of the total variability in the data, while the next four PCs, accounted for only 2.6\% of variability. Detailed variability results are shown in Table \ref{tab:PCs2}.

\begin{table}[htbp]
\centering
\caption{Explanation of Variability by Various Principal Components (PCs)}
\begin{tabular}{lcccccc}
\hline
  & PC1 & PC2 & PC3 & PC4 & PC5 & PC6 \\
\hline
Standard deviation& 191.24 & 23.63 & 19.22 & 14.78 & 11.42 & 10.17 \\
Proportion of Variance& 0.916 & 0.026 & 0.014 & 0.006 & 0.004 & 0.002 \\
Cumulative Proportion & 0.916 & 0.942 & 0.956 & 0.962 & 0.966 & 0.968 \\
\hline
\end{tabular}
\label{tab:PCs2}
\end{table}

\begin{figure*}[ht]
\centering
\includegraphics[width=0.98\textwidth]{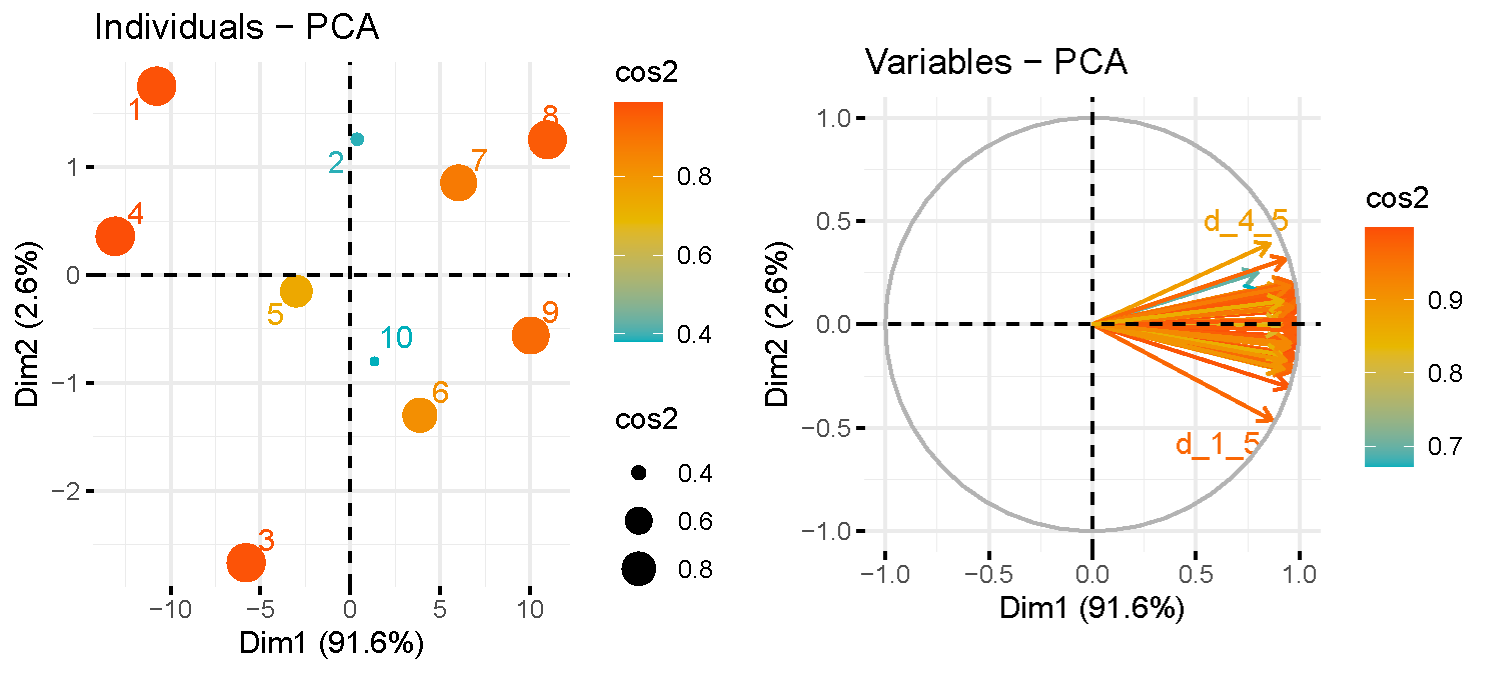}
\caption{PCA Analysis. In the left panel, the circles represent individual prawns with their dataset ID shown next to them. Their positions on the plot are determined by their scores on PC1 (Dim1) and PC2 (Dim2). The cos2 values measure the quality of representation of the individuals by the principal components. The right panel illustrates how different variables (distances) contribute to the variation in shape among prawns with respect to the two main PCs. Here, each of the arrows shows one of the distances, e.g. d\_1\_5 designates the distance between landmark 1 and 5, and how it relates to the two PCs.}
\label{fig_3}
\end{figure*}

The results for the first two PCs are also shown in Fig. \ref{fig_3}. In the left panel, the circles represent individual prawns with their dataset ID shown next to them. Their positions on the plot are determined by their scores on PC1 (Dim1) and PC2 (Dim2). The cos2 values measure the quality of representation of the individuals by the principal components.

PC1 accounted for the highest proportion of the variation (91.6\%) among the landmarks and was used as a representation of the overall size of the prawns. The positive loadings of PC1 indicate that an increase in PC1 scores corresponds to an increase in the distance between landmarks 1 and 3, which represents the total length of the prawn. PC1 can be used as an indicator of overall size, and the higher variance accounted for by PC1 may be due to the presence of a wide range of sizes in the samples.

PC2 (Dim2) explained 2.6\% of the total variation among landmarks and represents the proportionate body width/shape of the prawns. The extremes on the PC2 axis on Fig. \ref{fig_3}(left) represent a very thin/elongated body shape at the low end and a thick/fatty shape at the high end. 

Figure \ref{fig_3})(right) illustrates how different variables (distances) contribute to the variation in shape among prawns with respect to the two main PCs, i.e. PC1 and PC2. Here, each of the arrows shows one of the distances, e.g. d\_1\_5 designates the distance between landmark 1 and 5, and how it relates to the two PCs.

In conclusion, our PCA analysis of the predicted landmark data revealed that PC1 represents the overall size and PC2 represents the proportionate body width/shape of the prawns. These findings provide important insights into our targeted morphometric characteristics of prawns and lay a foundation for further research on the genetic and environmental factors that affect prawn morphology \cite{Hasan2022GeneticInvestigations}.

\section{Discussion} \label{secdisc}

The accurate estimation of weight and other morphological features of individuals being farmed in aquaculture is crucial data that when acquired on industrial scales can improve crop manage- ment, support decision-making, and be embedded in product grading and processing activities. Traditional methods of monitoring and acquiring morphological data from aquaculture animals are laborious, hence the need for developing automated methods using computer vision and deep learning.  In this paper, we proposed a novel approach that uses a Kronecker product-based feature extraction module and a landmark localization module. We utilized these efficient modules to extract important landmarks on the prawn body in a highly accurate, fast, and efficient way to perform accurate weight estimation and morphometric analysis of prawns from images. Our work is the first that applies the Kronecker product operation for feature extraction in morphometrics analysis and uses a localization network for landmark detection in prawn images.

Our deep learning-based image processing approach can be a valuable tool for improving the efficiency and profitability of aquaculture and fisheries operations, while also promoting sustainable practices and minimizing environmental impacts. Our approach helps researchers and managers obtain accurate measurements of prawn size, weight, and other morphological
features without invasive or time-consuming manual methods. This allows potentially for more efficient monitoring of populations, aiding in informed decisions about stocking densities, feeding regimes, and harvesting schedules. Noninvasive and accurate weight data can also provide critical information for managing prawn populations and predicting future yields, enhancing decision support.

Our approach has several strengths. First, it is fully automated, which saves time and reduces manual labour. Secondly, it is accurate and robust, enabling reliable prawn weight estimation even in challenging environments. Thirdly, our use of Deep Learning techniques to extract features from prawn images is a promising area of research that has shown great potential for morphometric analyses.

However, our approach also has limitations. It heavily relies on the quality of input images, and low-quality images may not provide enough information for accurate prawn weight estimation. Additionally, large-scale annotated datasets are needed for training our Deep Learning architecture, which may be difficult to obtain in some situations.

In future work, we aim to address these limitations by exploring new approaches to handle low-quality images and developing efficient and automated methods for acquiring and annotating large-scale datasets. Furthermore, we will investigate the application of our approach to other species of aquatic animals and explore the potential for using it for other applications such as automated disease diagnosis and monitoring. We believe that our work can significantly contribute to prawn aquaculture management and production, opening up new opportunities for Deep Learning applications in aquatic animal research and aquaculture engineering.

\section{Conclusions}  \label{secconc}

In this paper, we have proposed a novel deep learning approach for automated morphometric analyses and weight estimation of prawns from images. Our approach is efficient, accurate, and has the potential to improve the efficiency and profitability of aquaculture and fisheries operations. We believe that our work can significantly contribute to prawn aquaculture management and production, opening up new opportunities for deep learning applications in aquatic animal research and aquaculture engineering.

\subsection{CO2 Emission Related to Experiments}
Experiments were conducted using a private infrastructure, which has a carbon efficiency of 0.432 kgCO$_2$eq/kWh. A cumulative of 500 hours of computation was performed on the hardware of type RTX 2080 Ti (TDP of 250W).
Total emissions are estimated to be 54 kgCO$_2$eq of which 0 percents were directly offset.
Estimations were conducted using the \href{https://mlco2.github.io/impact#compute}{MachineLearning Impact calculator} presented in \cite{Lacoste2019QuantifyingLearning}.

More in detail, in \cref{tab:results}, we compare our proposed  model with a  ResNet-18 for landmark detection.  We find that our model reduces both training time and carbon emissions by 25\%. The ResNet-18 takes about 20 hours and emits 2.16 kgCO$_2$eq, while our model takes about 15 hours and emits 1.62 kgCO$_2$eq. Carbon emissions are a major concern for training large deep-learning models. Therefore, we believe that our method is a small step towards more efficient and eco-friendly models.

\section*{Acknowledgement}
This research is supported by the Australian Research Training Program (RTP) Scholarship and Food Agility HDR Top-Up Scholarship. D. Jerry and M. Rahimi Azghadi acknowledge the Australian Research Council through their Industrial Transformation Research Hub program.  
 
\section*{Additional Information}
\textbf{Competing interests} The authors declare no competing interests.\\
\textbf{Ethical approval} This work was conducted with the approval of 

\ifCLASSOPTIONcaptionsoff
\newpage
\fi

\bibliographystyle{IEEEtran}
\bibliography{references}

\begin{thebibliography}{10}
\providecommand{\url}[1]{#1}
\csname url@samestyle\endcsname
\providecommand{\newblock}{\relax}
\providecommand{\bibinfo}[2]{#2}
\providecommand{\BIBentrySTDinterwordspacing}{\spaceskip=0pt\relax}
\providecommand{\BIBentryALTinterwordstretchfactor}{4}
\providecommand{\BIBentryALTinterwordspacing}{\spaceskip=\fontdimen2\font plus
\BIBentryALTinterwordstretchfactor\fontdimen3\font minus
  \fontdimen4\font\relax}
\providecommand{\BIBforeignlanguage}[2]{{%
\expandafter\ifx\csname l@#1\endcsname\relax
\typeout{** WARNING: IEEEtran.bst: No hyphenation pattern has been}%
\typeout{** loaded for the language `#1'. Using the pattern for}%
\typeout{** the default language instead.}%
\else
\language=\csname l@#1\endcsname
\fi
#2}}
\providecommand{\BIBdecl}{\relax}
\BIBdecl

\bibitem{Sang2020GenotypeMonodon}
N.~V. Sang, N.~T. Luan, N.~V. Hao, T.~V. Nhien, N.~T. Vu, and N.~H. Nguyen,
  ``{Genotype by environment interaction for survival and harvest body weight
  between recirculating tank system and pond culture in Penaeus monodon},''
  \emph{Aquaculture}, vol. 525, p. 735278, 8 2020.

\bibitem{Boyd2022PerspectivesReview}
C.~E. Boyd, R.~P. Davis, and A.~A. McNevin, ``{Perspectives on the mangrove
  conundrum, land use, and benefits of yield intensification in farmed shrimp
  production: A review},'' 2022.

\bibitem{Mitteroecker2009AdvancesMorphometrics}
P.~Mitteroecker and P.~Gunz, ``{Advances in Geometric morphometrics},''
  \emph{Evolutionary Biology}, vol.~36, no.~2, 2009.

\bibitem{Jahanbakht2021}
\BIBentryALTinterwordspacing
M.~Jahanbakht, W.~Xiang, L.~Hanzo, and M.~R. Azghadi, ``{Internet of Underwater
  Things and Big Marine Data Analytics - A Comprehensive Survey},'' \emph{IEEE
  Communications Surveys and Tutorials}, vol.~23, no.~2, pp. 904--956, 2021.
  [Online]. Available: \url{https://ieeexplore.ieee.org/document/9328873/}
\BIBentrySTDinterwordspacing

\bibitem{Konovalov2019}
D.~A. Konovalov, A.~Saleh, D.~B. Efremova, J.~A. Domingos, and D.~R. Jerry,
  ``{Automatic Weight Estimation of Harvested Fish from Images},'' in
  \emph{2019 Digital Image Computing: Techniques and Applications, DICTA
  2019}.\hskip 1em plus 0.5em minus 0.4em\relax Institute of Electrical and
  Electronics Engineers Inc., 12 2019.

\bibitem{saleh2022a}
\BIBentryALTinterwordspacing
A.~Saleh, M.~Sheaves, and M.~Rahimi~Azghadi, ``{Computer vision and deep
  learning for fish classification in underwater habitats: A survey},''
  \emph{Fish and Fisheries}, vol.~23, no.~4, pp. 977--999, 7 2022. [Online].
  Available: \url{https://onlinelibrary.wiley.com/doi/10.1111/faf.12666}
\BIBentrySTDinterwordspacing

\bibitem{Setiawan2022ShrimpApproach}
A.~Setiawan, H.~Hadiyanto, and C.~E. Widodo, ``{Shrimp Body Weight Estimation
  in Aquaculture Ponds Using Morphometric Features Based on Underwater Image
  Analysis and Machine Learning Approach},'' \emph{Revue d'Intelligence
  Artificielle}, vol.~36, no.~6, pp. 905--912, 12 2022.

\bibitem{Vo2021OverviewVision}
\BIBentryALTinterwordspacing
T.~T.~E. Vo, H.~Ko, J.~H. Huh, and Y.~Kim, ``{Overview of Smart Aquaculture
  System: Focusing on Applications of Machine Learning and Computer Vision},''
  \emph{Electronics 2021, Vol. 10, Page 2882}, vol.~10, no.~22, p. 2882, 11
  2021. [Online]. Available: \url{https://www.mdpi.com/2079-9292/10/22/2882/htm
  https://www.mdpi.com/2079-9292/10/22/2882}
\BIBentrySTDinterwordspacing

\bibitem{Laradji2021}
\BIBentryALTinterwordspacing
I.~H. Laradji, A.~Saleh, P.~Rodriguez, D.~Nowrouzezahrai, M.~R. Azghadi, and
  D.~Vazquez, ``{Weakly supervised underwater fish segmentation using affinity
  LCFCN.}'' \emph{Scientific reports}, vol.~11, no.~1, p. 17379, 12 2021.
  [Online]. Available: \url{https://www.nature.com/articles/s41598-021-96610-2
  http://www.ncbi.nlm.nih.gov/pubmed/34462458
  http://www.pubmedcentral.nih.gov/articlerender.fcgi?artid=PMC8405733}
\BIBentrySTDinterwordspacing

\bibitem{Saleh2020a}
\BIBentryALTinterwordspacing
A.~Saleh, I.~H. Laradji, D.~A. Konovalov, M.~Bradley, D.~Vazquez, and
  M.~Sheaves, ``{A realistic fish-habitat dataset to evaluate algorithms for
  underwater visual analysis},'' \emph{Scientific Reports}, vol.~10, no.~1, p.
  14671, 12 2020. [Online]. Available:
  \url{https://www.nature.com/articles/s41598-020-71639-x}
\BIBentrySTDinterwordspacing

\bibitem{Devi2021KroneckerProduct}
J.~V. Devi, S.~G. Deo, and R.~Khandeparkar, ``{Kronecker Product},'' in
  \emph{Linear Algebra to Differential Equations}, 2021.

\bibitem{hung2014}
\BIBentryALTinterwordspacing
D.~Hung, N.~H. Nguyen, D.~A. Hurwood, and P.~B. Mather, ``{Quantitative genetic
  parameters for body traits at different ages in a cultured stock of giant
  freshwater prawn (Macrobrachium rosenbergii) selected for fast growth},''
  \emph{Marine and Freshwater Research}, vol.~65, no.~3, p. 198, 2014.
  [Online]. Available: \url{http://www.publish.csiro.au/?paper=MF13111}
\BIBentrySTDinterwordspacing

\bibitem{Lin2014b}
T.~Y. Lin, M.~Maire, S.~Belongie, J.~Hays, P.~Perona, D.~Ramanan,
  P.~Doll{\'{a}}r, and C.~L. Zitnick, ``{Microsoft COCO: Common objects in
  context},'' in \emph{Lecture Notes in Computer Science (including subseries
  Lecture Notes in Artificial Intelligence and Lecture Notes in
  Bioinformatics)}, 2014.

\bibitem{Ronneberger2015}
O.~Ronneberger, P.~Fischer, and T.~Brox, ``{U-net: Convolutional networks for
  biomedical image segmentation},'' in \emph{International Conference on
  Medical image computing and computer-assisted intervention}, 2015, pp.
  234--241.

\bibitem{He2015ResNet}
K.~He, X.~Zhang, S.~Ren, and J.~Sun, ``{Deep Residual Learning for Image
  Recognition},'' \emph{Computer Vision and Pattern Recognition (CVPR)}, 2015.

\bibitem{Zhang2018b}
X.~Zhang, X.~Zhou, M.~Lin, and J.~Sun, ``{ShuffleNet: An Extremely Efficient
  Convolutional Neural Network for Mobile Devices},'' in \emph{Proceedings of
  the IEEE Computer Society Conference on Computer Vision and Pattern
  Recognition}, 2018, pp. 6848--6856.

\bibitem{MobileNetV2}
M.~Sandler, A.~G. Howard, M.~Zhu, A.~Zhmoginov, and L.-C. Chen, ``{Inverted
  Residuals and Linear Bottlenecks: Mobile Networks for Classification,
  Detection and Segmentation},'' \emph{CoRR}, vol. abs/1801.0, 2018.

\bibitem{Iandola2016}
\BIBentryALTinterwordspacing
F.~N. Iandola, S.~Han, M.~W. Moskewicz, K.~Ashraf, W.~J. Dally, and K.~Keutzer,
  ``{SqueezeNet: AlexNet-level accuracy with 50x fewer parameters and <0.5MB
  model size},'' \emph{Computer Vision and Pattern Recognition}, 2 2016.
  [Online]. Available: \url{http://arxiv.org/abs/1602.07360}
\BIBentrySTDinterwordspacing

\bibitem{Kingma2014Adam:Optimization}
D.~P. Kingma and J.~Ba, ``{Adam: A method for stochastic optimization},''
  \emph{arXiv preprint arXiv:1412.6980}, 2014.

\bibitem{Paszke2019}
A.~Paszke, S.~Gross, F.~Massa, A.~Lerer, J.~Bradbury, G.~Chanan, T.~Killeen,
  Z.~Lin, N.~Gimelshein, L.~Antiga, A.~Desmaison, A.~K{\"{o}}pf, E.~Yang,
  Z.~DeVito, M.~Raison, A.~Tejani, S.~Chilamkurthy, B.~Steiner, L.~Fang,
  J.~Bai, and S.~Chintala, ``{PyTorch: An imperative style, high-performance
  deep learning library},'' in \emph{Advances in Neural Information Processing
  Systems}, 2019.

\bibitem{Konovalov2019b}
D.~A. Konovalov, A.~Saleh, D.~B. Efremova, J.~A. Domingos, and D.~R. Jerry,
  ``{Automatic weight estimation of harvested fish from images},'' in
  \emph{Digital Image Computing: Techniques and Applications (DICTA)}, 2019,
  pp. 1--7.

\bibitem{Bravata2020}
N.~Bravata, D.~Kelly, J.~Eickholt, J.~Bryan, S.~Miehls, and D.~Zielinski,
  ``{Applications of deep convolutional neural networks to predict length,
  circumference, and weight from mostly dewatered images of fish},''
  \emph{Ecology and Evolution}, 2020.

\bibitem{Hasan2022GeneticInvestigations}
\BIBentryALTinterwordspacing
M.~M. Hasan, P.~C. Thomson, H.~W. Raadsma, and M.~S. Khatkar, ``{Genetic
  analysis of digital image derived morphometric traits of black tiger shrimp
  (Penaeus monodon) by incorporating G × E investigations},'' \emph{Frontiers
  in Genetics}, vol.~13, 10 2022. [Online]. Available:
  \url{https://pubmed.ncbi.nlm.nih.gov/36338959/}
\BIBentrySTDinterwordspacing

\bibitem{Lacoste2019QuantifyingLearning}
\BIBentryALTinterwordspacing
A.~Lacoste, A.~Luccioni, V.~Schmidt, and T.~Dandres, ``{Quantifying the Carbon
  Emissions of Machine Learning},'' 10 2019. [Online]. Available:
  \url{/green-ai/publications/2019-11-lacoste-quantifying.html}
\BIBentrySTDinterwordspacing

\end{thebibliography}

\end{document}